\documentclass{article}
\usepackage{spconf,amsmath,graphicx}
\usepackage{graphicx}
\usepackage{amssymb}
\usepackage{booktabs}
\usepackage{todonotes}
\usepackage{multirow}
\usepackage{comment}
\usepackage{color, colortbl}
\definecolor{babyblue}{rgb}{0.54, 0.81, 0.94}
\newcommand*{\belowrulesepcolor}[1]{%
  \noalign{%
    \kern-\belowrulesep 
    \begingroup 
      \color{#1}%
      \hrule height\belowrulesep 
    \endgroup 
  }%
} 
\newcommand*{\aboverulesepcolor}[1]{%
  \noalign{%
    \begingroup 
      \color{#1}%
      \hrule height\aboverulesep 
    \endgroup 
    \kern-\aboverulesep 
  }%
} 
\usepackage{tabularx}
\usepackage{booktabs}
\usepackage{multirow}
\usepackage{array}
\usepackage{etoolbox}
\patchcmd{\thebibliography}
  {\settowidth}
  {\setlength{\itemsep}{0pt plus 0.1pt}\settowidth}
  {}{}
\apptocmd{\thebibliography}
  {\small}
  {}{}
\usepackage{amsmath}

\newcolumntype{H}{>{\setbox0=\hbox\bgroup}c<{\egroup}@{}}
\usepackage{xcolor,colortbl}
\usepackage[linesnumbered,ruled,vlined]{algorithm2e}
\usepackage{algpseudocode}
\usepackage{subcaption}
\usepackage{caption} 
\usepackage{hyperref}
\usepackage{url}
\usepackage[font=normal,labelfont=bf]{caption}
\usepackage{graphicx}
\usepackage{color, colortbl}
\usepackage{xcolor,colortbl}
\usepackage{float}
\newcommand{\R}{\mathbb{R}}
\expandafter\def\expandafter\normalsize\expandafter{%
    \normalsize%
    \setlength\abovedisplayskip{2pt}%
    \setlength\belowdisplayskip{6pt}%
    \setlength\abovedisplayshortskip{-8pt}%
    \setlength\belowdisplayshortskip{2pt}%
}
\title{Deep OC-SORT: Multi-Pedestrian Tracking by Adaptive Re-Identification}
\name{
Gerard Maggiolino$~^{*}\thanks{*: indicates equal contribution}$, 
Adnan Ahmad$~^{*}$, 
Jinkun Cao, 
Kris Kitani}
\address{Carnegie Mellon University}

\begin{document}
%
\maketitle

\begin{abstract}
Motion-based association for Multi-Object Tracking (MOT) has recently re-achieved prominence with the rise of powerful object detectors. Despite this, little work has been done to incorporate appearance cues beyond simple heuristic models that lack robustness to feature degradation. In this paper, we propose a novel way to leverage objects' appearances to adaptively integrate appearance matching into existing high-performance motion-based methods. Building upon the pure motion-based method OC-SORT, we achieve \textbf{1st place on MOT20} and \textbf{2nd place on MOT17} with 63.9 and 64.9 HOTA, respectively. We also achieve 61.3 HOTA on the challenging DanceTrack benchmark as a new state-of-the-art even compared to more heavily-designed methods. The code and models are available at \url{https://github.com/GerardMaggiolino/Deep-OC-SORT}.
\end{abstract}

\section{Introduction}
\label{sec:intro}
\vspace{-0.2cm}
With the success of advanced object detectors and motion-based association algorithms~\cite{bytetrack, cao2022observation, botsort}, the effective integration of visual appearance with motion-based matching remains relatively under-explored beyond simple moving average models~\cite{botsort, strongsort}. In this work, we start from a recent pure motion-based tracking algorithm OC-SORT~\cite{cao2022observation} and improve the tracking robustness by incorporating visual appearance with a novel approach. Bounding box-level visual features from strong embedding models still contain significant noise due to occlusion, motion blur, or objects of similar appearance. We propose a dynamic and adaptive heuristic-based model to incorporate the visual appearance with motion-based cues in a single stage for object association. Without fine-grained semantics, such as instance segmentation, we improve the accuracy of using visual comparison among objects for association. In addition to the contribution of more effectively adding appearance cues to motion-based object association, we integrate camera motion compensation, boosting performance by complementing the object-centric motion model. Our method provides a new and effective baseline model for future works. It sets a new state-of-the-art among all published works on MOT17~\cite{milan2016mot16}, MOT20~\cite{dendorfer2020mot20}, and DanceTrack~\cite{sun2021dancetrack} benchmarks. As our focus is to introduce visual appearance to OC-SORT, we name our method Deep OC-SORT. We note that the adaptive way we incorporate visual appearance with the motion-based method is newly designed, instead of a straightforward adaptation of what DeepSORT~\cite{deepsort} does upon SORT~\cite{bewley2016simple}.
\vspace{-0.3cm}

\section{Related Works}
\label{sec:related_works}
\vspace{-0.3cm}
\noindent \textbf{Motion-based Multi-Object Tracking.}
Given the rapid improvement of object detectors, many modern end-to-end MOT models still underperform against classic motion model-based tracking algorithms. The Kalman filter~\cite{kalman1960contributions} is the foundation of the most famous line of tracking-by-detection methods. Among this line of work, SORT~\cite{bewley2016simple} uses a linear motion assumption to associate tracks by IoU. ByteTrack~\cite{bytetrack} is recently proposed to fix missing predictions by using low-confidence candidates in association, achieving good performance by balancing the detection quality and tracking confidence. More recently, OC-SORT~\cite{cao2022observation} improves the robustness of tracking in non-linear motion scenarios and relieves the influence from object occlusion or disappearance by more heavily relying on detections directly. 

\noindent \textbf{Appearance-based Multi-Object Tracking.}
Visual identification is a straightforward cue to associate targets over time. DeepSORT~\cite{deepsort} is one of the earliest to use deep visual features for object association. Since then more methods ~\cite{zhang2021fairmot,pang2021quasi,cao2022track} have improved upon integrating visual information by training discriminative appearance models in an end-to-end manner. More recently, the rise of transformers~\cite{vaswani2017attention} has started another wave of using appearance for multi-object tracking, where the task of object association is modeled as a query matching problem~\cite{zeng2021motr,sun2020transtrack,meinhardt2021trackformer,cao2022track}. However, appearance-based methods are observed to be less effective when the objects of interest have similar appearance~\cite{sun2021dancetrack} or are occluded~\cite{dendorfer2020mot20,sun2021dancetrack}. Despite having more complicated architectures, these methods fail to outperform simple motion association algorithms that leverage strong detectors. Some recent attempts to add appearance cues~\cite{botsort,strongsort} to motion-based methods use simple moving averages for appearance embedding updates, achieving moderate success. 

\section{Methods}
\label{sec:methods}
\begin{figure}[h!]
    \centering
    \includegraphics[width=0.9\linewidth]{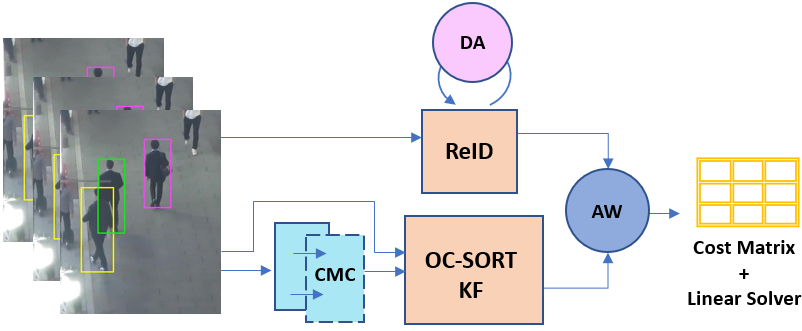}
    \caption{Illustration of Deep OC-SORT.}
    \label{fig:cloth_characteristic2}
\end{figure}

\vspace{-0.2cm}
In this section, we describe the three modules of Deep OC-SORT: Camera Motion Compensation (CMC), Dynamic Appearance (DA) and Adaptive Weighting (AW). The algorithm pipeline is illustrated in Figure~\ref{fig:cloth_characteristic2}. 

\subsection{Preliminary: OC-SORT}
Our work is built upon the recent Kalman-filter-based tracking algorithm OC-SORT~\cite{cao2022observation}, which is an extension of SORT~\cite{bewley2016simple}. SORT relies on the linear motion assumption of object tracking and leverages the Kalman filter to associate predictions from an object detector with the position estimates from the motion model by IoU. When the video frame rate is high, the linear motion assumption can be effective for object displacement on adjacent video frames. However, when tracking targets disappear under occlusion, the missed measurements during Kalman filter updates compound error quadratically over time in the Kalman filter's parameters. OC-SORT proposes three modules to help resolve the motion-model based error: OCM (observation-centric momentum), OCR (observation-centric recovery), and OOS (observation-centric online smoothing). We invite the reader to refer to its paper~\cite{cao2022observation} for details.
We inherit the overall pipeline of OC-SORT, including the Hungarian algorithm to associate matches from a cost matrix.

\begin{table*}[t]
\centering
\caption{Results on MOT17-test and MOT20-test. Methods in the blue blocks share the same detections.}
\setlength{\tabcolsep}{7pt}
\scriptsize
\begin{tabular}{ l | p{20px}p{20px}p{20px}p{27px} p{27px}p{22px}p{22px}p{20px}p{20px}}
\toprule
\belowrulesepcolor{gray!20} 
\rowcolor{gray!20} \multicolumn{10}{c}{\textbf{MOT17}}
\\
\aboverulesepcolor{gray!20} 
\midrule
Tracker &  HOTA$\uparrow$ & MOTA$\uparrow$ & IDF1$\uparrow$ &  FP({\footnotesize $10^4$})$\downarrow$ & FN({\footnotesize $10^4$})$\downarrow$ & IDs$\downarrow$ & Frag$\downarrow$ & AssA$\uparrow$ & AssR$\uparrow$ \\
\midrule
FairMOT~\cite{zhang2021fairmot} & 59.3 & 73.7 & 72.3  & 2.75 & 11.7 & 3,303 & 8,073 & 58.0 & 63.6\\
TransCt~\cite{transcenter} & 54.5 & 73.2 & 62.2  & 2.31 & 12.4 & 4,614 & 9,519 & 49.7 & 54.2 \\
TransTrk~\cite{sun2020transtrack} & 54.1 & 75.2 & 63.5  & 5.02 & 8.64 & 3,603 & 4,872 & 47.9 & 57.1 \\
GRTU~\cite{grtu} & 62.0 & 74.9 & 75.0  & 3.20 & 10.8 & 1,812 & \textbf{1,824} & 62.1 & 65.8\\ 
QDTrack~\cite{pang2021quasi} & 53.9 & 68.7 & 66.3  & 2.66 & 14.66 & 3,378 & 8,091 & 52.7 & 57.2\\
MOTR~\cite{zeng2021motr} & 57.2 &71.9 & 68.4 & 2.11 & 13.6 & 2,115 & 3,897 & 55.8 & 59.2  \\
TransMOT~\cite{chu2021transmot} & 61.7 & 76.7 & 75.1 & 3.62 & 9.32 & 2,346 & 7,719 & 59.9 & 66.5 \\
\midrule
\belowrulesepcolor{babyblue!20} 
\rowcolor{babyblue!20}ByteTrack~\cite{bytetrack} & 63.1 & \textbf{80.3} & 77.3 & 2.55 &  \textbf{8.37} & 2,196 & 2,277 & 62.0 & 68.2\\
\rowcolor{babyblue!20}OC-SORT~\cite{cao2022observation} & 63.2 & 78.0 & 77.5 & \textbf{1.51} & 10.8 & 1,950 & 2,040 & 63.2 & 67.5\\
\rowcolor{babyblue!20}StrongSORT~\cite{strongsort} & 63.5 & 78.3 & 78.5 & - & - & 1,446 & - & 63.7 & -\\ 
\rowcolor{babyblue!20}*StrongSORT++~\cite{strongsort} & 64.4 & 79.6 & 79.5 & 2.79 & 8.62 & 1,194 & \textbf{1,866} & 64.4 & \textbf{71.0}\\
\rowcolor{babyblue!20}Deep OC-SORT & \textbf{64.9} & 79.4 & \textbf{80.6} & 1.66 & 9.88 & \textbf{1,023} & 2,196 & \textbf{65.9} & 70.1\\ 
\aboverulesepcolor{babyblue!20}
\midrule
\belowrulesepcolor{gray!20} 
\rowcolor{gray!20} \multicolumn{10}{c}{\textbf{MOT20}}
\\
\aboverulesepcolor{gray!20} 
\midrule
Tracker &  HOTA$\uparrow$ & MOTA$\uparrow$ & IDF1$\uparrow$ &  FP({\footnotesize $10^4$})$\downarrow$ & FN({\footnotesize $10^4$})$\downarrow$ & IDs$\downarrow$ & Frag$\downarrow$ & AssA$\uparrow$ & AssR$\uparrow$ \\
\midrule
FairMOT~\cite{zhang2021fairmot} & 54.6 & 61.8 & 67.3 & 10.3 & 8.89 & 5,243 & 7,874 & 54.7 & 60.7 \\
Semi-TCL~\cite{li2021semitcl} & 55.3 & 65.2 & 70.1 & 6.12 & 11.5 & 4,139 & 8,508 & 56.3 & 60.9 \\
CSTrack~\cite{cstrack} & 54.0 & 66.6 & 68.6  & 2.54 & 14.4 & 3,196 & 7,632 &  54.0 & 57.6 \\
GSDT~\cite{gsdt} & 53.6 & 67.1 & 67.5  & 3.19 & 13.5 & 3,131 & 9,875 & 52.7 & 58.5 \\
TransMOT~\cite{chu2021transmot} & 61.9 & 77.5 & 75.2 & 3.42 & \textbf{8.08} & 1,615 & 2,421 & 60.1 & 66.3 \\
\midrule
\belowrulesepcolor{babyblue!20} 
\rowcolor{babyblue!20} ByteTrack~\cite{bytetrack} & 61.3 & \textbf{77.8} & 75.2 &  2.62 & 8.76 & 1,223 & 1,460 & 59.6 & 66.2\\
\rowcolor{babyblue!20}OC-SORT~\cite{cao2022observation} & 62.1 & 75.5 & 75.9 &  1.80 & 10.8& 913 & 1,198& 62.0 & 67.5 \\
\rowcolor{babyblue!20}StrongSORT~\cite{strongsort} & 61.5 & 72.2 & 75.9 & - & - & 1,066 & - & 63.2 &  \\ 
\rowcolor{babyblue!20}*StrongSORT++~\cite{strongsort} & 62.6 & 73.8 & 77.0 & \textbf{1.66} & 11.8 & \textbf{770} & \textbf{1,003} & 64.0 & 69.6 \\
\rowcolor{babyblue!20}Deep OC-SORT & \textbf{63.9} & 75.6 & \textbf{79.2} & 1.69 & 10.8 & 779 & 1,536 & \textbf{65.7} & \textbf{70.8}\\
\aboverulesepcolor{babyblue!20}
\bottomrule
\end{tabular}
\begin{center}\scriptsize
\vspace{-0.4cm}
\item[*]: StrongSORT++ requires offline post-processing while ByteTrack, OC-SORT, StrongSORT and Deep OC-SORT are for online multi-object tracking.
\end{center}
\label{table:mot17}
\vspace{-0.5cm}
\end{table*}

\begin{table}[hbt!]
\centering
\caption{Results on DanceTrack test set. Methods in the blue block use the same detections.}
\setlength{\tabcolsep}{7pt}
\scriptsize
\begin{tabular}{ l | p{2.2em}p{2em}p{2em}p{2.3em}p{2.3em}}
\toprule
Tracker & HOTA$\uparrow$ & DetA$\uparrow$ & AssA$\uparrow$ & MOTA$\uparrow$ & IDF1$\uparrow$\\
\midrule
CenterTrack~\cite{centertrack} & 41.8 & 78.1 & 22.6 & 86.8 & 35.7 \\
FairMOT~\cite{zhang2021fairmot} & 39.7 & 66.7 & 23.8 & 82.2 & 40.8\\
QDTrack~\cite{pang2021quasi} & 45.7 & 72.1 & 29.2 & 83.0 & 44.8\\
TransTrk\cite{sun2020transtrack} & 45.5 & 75.9 & 27.5 & 88.4 & 45.2\\
TraDes~\cite{trades} & 43.3 & 74.5 & 25.4 & 86.2 & 41.2 \\ 
MOTR~\cite{zeng2021motr} & 54.2 & 73.5 & 40.2 & 79.7 & 51.5\\
\midrule
\belowrulesepcolor{babyblue!20} 
\rowcolor{babyblue!20} SORT~\cite{bewley2016simple} & 47.9 & 72.0 & 31.2 & 91.8 & 50.8 \\
\aboverulesepcolor{babyblue!20} 
\rowcolor{babyblue!20}DeepSORT~\cite{deepsort} & 45.6 & 71.0 & 29.7 & 87.8 & 47.9\\
\rowcolor{babyblue!20}ByteTrack~\cite{bytetrack} & 47.3 & 71.6 & 31.4 & 89.5 & 52.5\\
\rowcolor{babyblue!20}OC-SORT~\cite{cao2022observation} & 55.1 & 80.3 & 38.3 & 92.0 & 54.6\\
\rowcolor{babyblue!20}*StrongSORT++~\cite{strongsort} & 55.6 & 80.7 & 38.6 & 91.1 & 55.2\\
\rowcolor{babyblue!20}Deep OC-SORT & \textbf{61.3} & \textbf{82.2} & \textbf{45.8} & \textbf{92.3} & \textbf{61.5}\\
\aboverulesepcolor{babyblue!20} 
\bottomrule
\end{tabular}
\begin{center}\scriptsize
\vspace{-0.4cm}
\item[*]: StrongSORT++ requires offline post-processing while others are for online tracking.
\end{center}
\label{table:dancetrack}
\vspace{-0.5cm}
\end{table}

\vspace{-2em}

\subsection{Camera Motion Compensation (CMC)}
As OC-SORT is highly dependent on the detection quality, we introduce CMC to more accurately localize objects from frame to frame in moving scenes. 
Given a scaled rotation matrix $M_t = s_tR_t$ and a translation $T_t$ where $M_t \in \R^{2\times2}$and $T_t \in \R^{2\times1}$,  we apply them to OC-SORT's three components respectively:
\begin{enumerate}
    \item \textbf{OOS + CMC.} The Kalman filter is updated from the linearly interpolated path, starting at the last known measurement. This last known measurement is comprised of $[x_c, y_c, a, s]$, with the first two entries as the center of the bounding box. The center of the bounding box is similarly transformed by $c \leftarrow M_tc + T_t $, so that the path is interpolated starting from the camera corrected measurement. 
    \item \textbf{OCM + CMC.} Let $p_1, p_2$ be the upper-left and lower-right corner points of a bounding box. OCM uses the last $\Delta t = 3$ bounding boxes to compute a bounding box angular velocity. At each timestep $t$, we apply the transformation $p_i \leftarrow M_tp_i + T_t $  to the bounding box. This goes from $t - \Delta t$ to timestep $t$ during OCM. 
    \item \textbf{OCR + CMC.} For the last-seen bounding box position in OCR, at each timestep $t$, we apply $p_i \leftarrow M_tp_i + T_t $ to adjust its position under CMC.
\end{enumerate}

For OC-SORT, the Kalman state is $\mathbf{x} = [x_c,  y_c, a, s, \dot{x_c}, \dot{y_c}, \dot{a}]$. We apply CMC to correct the Kalman state:
\begin{equation}
\left\{ 
\begin{aligned}
\bold{x}[0:2] &\leftarrow M_t \bold{x}[0:2] + T_t,\\
\bold{x}[4:6] &\leftarrow M_t \bold{x}[4:6], \\
P[0:2, 0:2] &\leftarrow M_t P[0:2, 0:2] M_t^T,\\
P[4:6,4:6] &\leftarrow M_t P[4:6, 4:6] M_t^T.
\end{aligned}
\right. 
\end{equation}

We note that we could apply the \textit{scale} of the CMC transform to the area $a$, or approximate rotation to change the aspect ratio $s$. However, in contrast to the center point, the enclosing bounding box of a rotated object is not approximated linearly and requires a fine-grained mask of the enclosed object. While the approximation works well with OCM and OCR, the Kalman filter is empirically more sensitive to approximate changes. We apply this CMC update before the Kalman extrapolation step so that the \textit{prediction} stage is from the CMC-corrected states. 

\subsection{Dynamic Appearance} 
In previous work~\cite{botsort, strongsort}, the deep visual embedding used to describe a tracklet is given by an Exponential Moving Average (EMA) of the deep detection embeddings frame by frame. This requires a weighting factor $\alpha$ to adjust the ratio of the visual embedding from historical and current time steps. We propose to modify the $\alpha$ of the EMA on a per-frame basis, depending on the detector confidence. This flexible $\alpha$ allows selectively incorporating appearance information into a track's model only in high-quality situations. 

\begin{table*}[!htb]
\centering
\caption{Ablation study on MOT17-val, MOT20-val and DanceTrack-val set.}
\setlength{\tabcolsep}{7pt}
\scriptsize
\begin{tabular}{ cccc | ccc | ccc | ccc}
\toprule
\multicolumn{4}{c|}{} & \multicolumn{3}{c|}{MOT17-val} & \multicolumn{3}{c|}{MOT20-val} & \multicolumn{3}{c}{DanceTrack-val}\\ 
\midrule
Appr. & DA & CMC & AW & HOTA$\uparrow$ &  AssA$\uparrow$ & IDF1$\uparrow$ & HOTA$\uparrow$ &  AssA$\uparrow$ & IDF1$\uparrow$ & HOTA$\uparrow$ &  AssA$\uparrow$ & IDF1$\uparrow$\\ 
\midrule
& & & & 68.13 & 70.06 & 79.52 & 58.35 & 56.11 & 74.77 & 53.07 & 35.93 & 52.43\\
& & \checkmark & & 69.66 & 72.72 & 82.44 & 58.33 & 56.12 & 74.63 & 53.57 & 36.68 & 53.30\\
\checkmark &  & & & 68.59 & 70.63 & 80.18 & 59.10 & 57.47 & 75.71 & 58.03 & 42.37 & 57.73\\
\checkmark & \checkmark & & & 68.65 & 70.85 &  80.45 & 59.16 & 57.60 & 75.87 & 58.36 & 43.00 & 58.17\\ 
\checkmark & \checkmark & \checkmark & & 69.80 & 72.86 & 82.56 & 59.35 & 58.00 & 76.11 & 58.46 & 43.33 & 58.83\\
\checkmark & \checkmark & \checkmark & \checkmark & \textbf{70.20} & \textbf{73.46} & \textbf{82.78} & \textbf{59.45} & \textbf{58.16} & \textbf{76.30} & \textbf{58.53} & \textbf{43.41} & \textbf{59.06}\\
\bottomrule
\end{tabular}
\label{table:ablation}
\vspace{-0.5cm}
\end{table*}

We use low detector confidence as a proxy to recognize image degradation due to occlusion or blur, allowing us to reject corrupted embeddings. Let $\mathbf{e}_t$ be the tracklet's appearance embedding at time $t$. The standard EMA is

\begin{equation}
    \mathbf{e}_t = \alpha \mathbf{e}_{t -1} + (1 - \alpha) \mathbf{e}^{\text{new}},
\end{equation}
where $\mathbf{e}^\text{new}$ is the appearance of the matched detection being added to the model. We propose replacing $\alpha$ with a changing $\alpha_t$ defined as
\vspace{-0.3cm}
\begin{equation}
    \alpha_t = \alpha_f + (1 - \alpha_f) (1 - \frac{s_\text{det} - \sigma}{1 - \sigma}),
\end{equation}
where $s_\text{det}$ is the detector confidence, and $\sigma$ is a detection confidence threshold to filter noisy detections, a common practice of previous works~\cite{bewley2016simple,cao2022observation,bytetrack,strongsort}. We set the fixed value $\alpha_f = 0.95$.  
The detector prediction provides $s_{\text{det}}$, controlling the dynamic operation. 
With $s_\text{det} = \sigma$, we have $\alpha_t = 1$, so that the new appearance embedding is totally ignored. 
In contrast, $s_\text{det} = 1$ implies $\alpha_t = \alpha_f$, and $\mathbf{e}^\text{new}$ is maximally added to the update of tracklet visual embedding. The value scales linearly with detector confidence. 
The operation to generate the dynamic appearance introduces no new hyper-parameters to the standard EMA.

\subsection{Adaptive Weighting} 
Our Adaptive Weighting increases the weight of appearance features depending on the discriminativeness of appearance embeddings.  Using standard cosine similarity across track and box embeddings results in an $M \times N$ appearance cost matrix, $A_c$ where $M$ and $N$ are the numbers of tracks and detections respectively. $A_c[m,n]$ indicates the entry at the intersection of the $m$-th row and the $n$-th column. This is typically combined with the IoU cost matrix $I_c$ as $C = I_c + a_w A_c$, with a linear sum assignment minimizing cost over $-C$. 

We propose to boost individual track-box scores based on discriminativeness, adding $w_b(m, n)$ to the global $a_w$. Let $\tau_m$ represent a track and $d_n$ represent a detection. When $\tau_m$ has a high similarity score to only one box (included in the row $A_{c}[m,:]$), we increase appearance weight over row $A_{c}[m,:]$. The same operation is applied to the columns of $A_c$ if a detection $d_n$ is associated discriminatively with only one track. We use $z_{\text{diff}}$ to measure the discriminativeness of box-track pairs, which is defined as the difference between the highest and second-highest values at a row or a column:
\begin{equation}
    \begin{aligned}
        z_{\text{diff}}^{\text{det}}(A_c, n) &= \text{min}(\max_{i} A_c[i,n] - \max_{j \neq i} A_c[j,n], \epsilon),\\
        z_{\text{diff}}^{\text{track}}(A_c, m) &= \text{min}(\max_{i} A_c[m,i] - \max_{j \neq i} A_c[m,j], \epsilon),
    \end{aligned}
\end{equation}
where $\epsilon$ is a hyper-parameter to cap the boost where there's a large difference in appearance cost between the first and second best matches.
Then, we derive the weighting factor as 
\begin{equation}
    w_b(m, n) = \left[z_{\text{diff}}^{\text{track}}(A_c, m) + z_{\text{diff}}^{\text{det}}(A_c, n)\right] / 2,
\end{equation}
which results in the final cost matrix $C$ as
\begin{equation}
    C[m, n] = \text{IoU}[m, n] +\left[a_w + w_b(m, n)\right] A_c[m, n].
\end{equation}
We choose to measure the discriminativeness based on only the first and second-highest scores rather than probability distribution metrics like KL divergence, as the spread of values between lower-scoring matches are irrelevant. A true positive appearance match is indicated by one high score having a large distance from the next best match.

\section{Experiments}
\label{sec:experiments}
\vspace{-0.2cm}
In this section, we provide experimental evidence to demonstrate the effectiveness of Deep SORT. We also analyze the influence of each module we introduce over OC-SORT~\cite{cao2022observation}. 

\noindent \textbf{Datasets and Metrics.} We conduct experiments on multiple datasets to ensure the generalizability of the proposed method. The investigated datasets include the popular pedestrian tracking datasets MOT17~\cite{milan2016mot16}, MOT20~\cite{dendorfer2020mot20}, and DanceTrack~\cite{sun2021dancetrack}. We follow the HOTA protocol\cite{luiten2021hota} for quantitative evaluation, which provides a more comprehensive measurement of the tracking quality. HOTA is the main metric we refer to for tracking performance. AssA is the metric to measure association accuracy and DetA is for detection accuracy. We also report the metrics in the classic CLEAR protocol~\cite{clearprotocol} for reference where MOTA indicates a overall performance of detection and tracking and IDF1 provides a measurement of association accuracy. 

\noindent \textbf{Implementations}
Our implementation is based on OC-SORT~\cite{cao2022observation,mmtrack2020}. We use the same YOLOX detector as recent works ~\cite{strongsort, botsort, cao2022track, bytetrack} to make a fair comparison of tracking performance. For Re-ID, we use SBS50 from the fast-reid~\cite{he2020fastreid} library. For CMC, we adopt the OpenCV contrib VidStab module to generate similarity transforms using feature point extraction, optical flow, and RANSAC, as previous works~\cite{botsort} choose. Across all experiments, we use a fixed $\alpha=0.95$ for Dynamic Appearance. For experiments on MOT17 and MOT20, we set $a_w = 0.75$ and $\epsilon = 0.5$ for Adaptive Weighting. 
We use $a_w = 1.25$ for DanceTrack, where we see appearance is more beneficial than IoU, and $\epsilon = 1.0$. 

\subsection{Benchmark Results}
\vspace{-0.2cm}
We conduct experiments on MOT17~\cite{milan2016mot16}, MOT20~\cite{dendorfer2020mot20}, and DanceTrack~\cite{sun2021dancetrack}. 
The results on MOT17-test and MOT20-test are shown in Table~\ref{table:mot17}. On MOT17-test, Deep OC-SORT achieves 64.9 HOTA, which outperforms all published methods and ranks 2nd on the leaderboard. 
On MOT20-test, Deep OC-SORT achieves 63.9 HOTA, ranking 1st on the leaderboard. Finally, on the most challenging dataset DanceTrack, where tracking algorithms usually suffer from heavy occlusion and frequent crossovers, our method achieves a new state-of-the-art among published methods as shown in Table~\ref{table:dancetrack}. 
On all three datasets, using the same detections, our method beats the existing comparisons including SORT~\cite{bewley2016simple}, DeepSORT~\cite{deepsort}, ByteTrack~\cite{bytetrack}, OC-SORT~\cite{cao2022observation}, and StrongSORT~\cite{strongsort}. 
Being online and without offline post-processing, our method still shows better association accuracy even compared to StrongSORT++, which is the offline version of StrongSORT, enhanced by offline post-processing of tracking trajectories. 
The benchmark results make strong evidence of the advanced performance of Deep OC-SORT.

\subsection{Ablation Study}
\vspace{-0.2cm}
To demonstrate the effectiveness of the proposed modules, we perform an ablation study on validation sets of MOT17, MOT20, and DanceTrack. With a baseline of OC-SORT, we describe performance with the addition of: Appearance Embedding with a fixed EMA (Appr.), Dynamic Appearance (DA), Camera Motion Compensation (CMC), and Adaptive Weighting (AW). The results are shown in Table~\ref{table:ablation}. 

We find that CMC improves performance on MOT17-val and DanceTrack-val sets while providing no improvements on MOT20-val, which is captured from static cameras. Appearance cues (Appr.) improve performance on all datasets across all metrics. Further applying Dynamic Appearance similarly boosts performance on all metrics, while adding no additional hyper-parameters and entirely negligible computation.  Finally, Adaptive Weighting provides yet another consistent improvement in performance across all metrics and datasets.

\section{Conclusion}
\vspace{-0.2cm}
In this work, we start from a motion-only multi-object tracking algorithm OC-SORT~\cite{cao2022observation} and propose a novel way of incorporating visual appearance. The proposed adaptive re-identification uses a weighted appearance similarity and compares across detection-track matches to create a blended visual cost. The extra camera motion compensation also provides benefits for tracking objects under moving cameras. We hope the implemented method can serve as a strong baseline for future studies with both motion and appearance cues taken into multi-object tracking.

\section{Acknowledgements}
\vspace{-0.2cm}
We would like to thank Rahul Nallamothu, Sam Pepose , Xiaochen Han and Chengxiang Yin from the Portal team at Meta for their useful advice and mentorship that helped shape the ideas for this work.

\bibliographystyle{IEEEbib}
\bibliography{refs}

\end{document}